# Real-time Rail Recognition Based on 3D Point Clouds

**Xinyi Yu[1], Weiqi He[1], Xuecheng Qian[1], Yang Yang[1], Linlin Ou[1]**

[1] Laboratory of Intelligent learning and Robotics, School of Information Engineering,
Zhejiang University of Technology, Zhejiang 310000, China

**Email**: 2111903075@zjut.edu.cn

**Abstract**

Accurate rail location is a crucial part in the railway support driving system for safety monitoring. LiDAR can obtain point clouds that carry 3D information for the railway environment, especially in darkness and terrible weather conditions. In this paper, a real-time rail recognition method based on 3D point clouds is proposed to solve the challenges, such as disorderly, uneven density and large volume of the point clouds. A voxel down-sampling method is first presented for density balanced of railway point clouds, and pyramid partition is designed to divide the 3D scanning area into the voxels with different volumes. Then, a feature encoding module is developed to find the nearest neighbor points and to aggregate their local geometric features for the center point. Finally, a multi-scale neural network is proposed to generate the prediction results of each voxel and the rail location. The experiments are conducted under 9 sequences of 3D point cloud data for the railway. The results show that the method has good performance in detecting straight, curved and other complex topologies rails.

## 1. Introduction

Railway transportation is an essential carrier of personnel flow, which is regarded as a comfortable, fast and safe way of transportation. Safety monitoring is a crucial part of railway transportation automation. At present, most routes still rely on the judgment of the drivers to ensure driving safety, such as identifying signal lights and obstacles on the rail. However, driver negligence and untimely reactions are inevitable and pose a significant challenge to train safety. Intelligent applications based on vehicular sensors, such as lane-keeping and auxiliary braking, have been developed to ensure safe driving for many years. Thus, it is an effective way to establish an auxiliary driving system of train based on vehicular sensors [1-4]. Recognizing rail location in front of the train is regarded as one of the essential functions of the auxiliary driving system. Based on this function, other security detection functions can be developed, such as signal lamp recognition[5], obstacle identification, collision prevention[6] and obstacle-free range detection[7].

The camera-based applications in driver-assistance systems have many advantages, such as rich image information and direct visual feedback for train drivers. In previous studies, rail extraction methods[2, 8] were based on 2D images captured by vehicle cameras. However, these methods require good lighting conditions and are less robust in rainy, foggy and night time environments.

Light Detection and Ranging (LiDAR) is an active sensor transmitting electromagnetic radiation and measuring backscattered signals. By measuring the attenuation of the incident light pulse, LiDAR reflects the properties of the objects within its range. Recently, LiDAR has been considered as an efficient tool for remote sensing and environment modeling[9]. It is widely used in autopilot cars and crewless aerial vehicles for generating an accurate 3D map of urban roads, dams, tunnels and large buildings[10, 11]. Meanwhile, LiDAR has made significant progress in automatic driving. This allows recognition of activities around the vehicle and plays a vital role in shaping decisions about the trajectory and characteristics of the motion. Furthermore, LiDAR has been introduced into Railway applications, such as railway measurement[12], gap measurement[13], infrastructure reconstruction[14] and tunnel mapping[15].

As an essential function, rail recognition requires timely operation and judgment of trains running at the high-speed. Therefore, the accuracy and discrimination speed of rail recognition are two critical problems. However, the point cloud data produced by LiDAR has two drawbacks. On one hand, a major challenge is increased data volume compared with images. The data must be processed in a limited amount of time, which requires an expansion of the algorithm processing capabilities. Previous researches[16-18] took down-sampling of point clouds to reduce time cost, but they pay little attention to the density and sparseness of point clouds in outdoor scenes. Meanwhile, point clouds are discrete and variable in density, which increases the difficulty for feature extraction of objects. On the other hand, LiDAR data is affected significantly by noise[19, 20], which deteriorates the signal-to-noise ratio. Although the noise is correlated with the LiDAR scanning distance and is occasionally handled as non-Gaussian, it causes some difficulty to recognize the rail with high accuracy. Additionally, the speed of the train, together with the relative motion of the rail, causes non-stationary behavior and increases extra difficulty.

The rail area, including a pair of equidistant tracks and rail bed, appears as an elongated strip in 3D point clouds collected by LiDAR. The point clouds contain the specific reflection intensity of the tracks. The intensity data of track surface shows as a slender line at the junction of train wheels. However, many kinds of rail topologies in the railway scene need to be located accurately. Thus, the following three problems have to be considered: (1) the number of point clouds in the railway scene is massive, and an appropriate down-sampling method is needed to reduce the calculation of the points feature extraction; (2) as the LiDAR scanning distance increase, the low density of point clouds and the weak geometric characteristics of the rail pose great difficulty for recognition; (3) for the points noise and complex types of track arrangement, straight, curve, fork and intersection, rail recognition methods need good robustness.

In this study, we propose a method to recognize rail real-timely with high accuracy in complex scenes, such as curved rails and crossed rials. A new 3D space division method named pyramid partition is proposed to make the points distribute evenly in each voxel. At the same time, the points discarded after voxel down-sampling only involve in the calculation during position encoding, which greatly save computation in next steps. Then the point cloud semantic segmentation neural network is introduced to adaptively learn the multi-scale features of points. It shows good generalization performance and reduces the impact of point cloud noise on the recognition accuracy. Finally, the proposed method is tested on a dataset collected with solid-state LiDAR in the real-world railway environment, including straight rail, curved rail and crossed rail. The major contributions of this study are listed as follows.

(1) This study proposes a density balanced voxel down-sampling method devised for the solid-state LiDAR scanning scope and solves the difficulties caused by the variable density of point clouds.

(2) This study proposes a point feature extraction (PFE) module, which can enrich the features of point voxel pairs by introducing the relative location and density information of points.

(3) An effective rail recognition neural network architect-ture is designed to achieve real-time rail recognition in various scenarios with good performance.

The structure of this paper is given as follows. Section 2 presents the related work of railway object recognition, including rail extraction based on image, object recognition based on mobile laser scanning (MLS) data, and point cloud semantic segmentation based on deep learning. Section 3 presents the details of the proposed method, including down-sampling with pyramid partition, point feature extraction module and multi-scale network structure based on 3D sparse convolution. Experiments and results are provided in Section 4 and conclusions are included in Section 5.

## 2. Related Work

The camera is widely used in railway systems because of its price advantage and direct visual feedback to humans. Images based methods include features matching and edge detection such as combining local grey value gradients[1], Inverse Projective Mapping (IPM) and sliding window method[21, 22]. Nassu[8] extracted rails by matching edge features to candidate rail patterns modeled as sequences of parabola segments. The robustness of long-distance rail matching algorithm is poor disturbing by rod objects, such as trees and telegraph poles. Conversely, Selver[2] used two-dimensional Gabor wavelets at multiple scales and directions as edge detectors on the partitioned video frame. Although the robustness of rail recognition is enhanced, it is difficult to adjust the parameters for various railway scenes and susceptible to mechanical jitter. Wang[22] extracted rails by matching the edge features of the IPM based image to the candidate rail patterns, which requires complex matrix solution and feature point selection. With the development of object detection[23, 24] based on deep learning, He[25] and Wang[26] regarded rail and obstacles as identification targets, and introduced convolutional neural networks (CNNs) to realize end-to-end identification without manually extracting track features. However, images only have 2D information which is the main reason for its poor robustness in complex railway environments and weather conditions. Compared with the camera, 3D information of LiDAR data greatly improves spatial details.

LiDAR is widely accepted for environmental perception in autonomous[27, 28]. The corresponding railway applications[29, 30] include detection, extraction, and modeling of objects and environments based on point clouds. Yang[31] implemented a moving window filtering operator for searching elevation jumps in points to extract rails. The random sample consensus (RANSAC) algorithm was used to extract rough planes in large-scale road sences[32]. It suppresses the influence of point cloud noise on feature segmentation with a high iterative calculation efficiency. However manually setting the plane threshold for modeling in advance leads to its poor robustness. Lou[33] used principal component analysis (PCA) method to cluster the rail points according to the mechanical structure characteristics of the relative between the rail and bed bulge. Rong[34] proposed a K-means clustering fused Region-Grow Fitting algorithm to calculate the direction and endpoints of the rail section. Although they randomly sampled point cloud seeds without setting thresholds, they are vulnerable to noise points and initial position of the seeds during clustering. Huang[35] combined the repeated least trimmed squares idea with the smoothing fairness function to filter noise and fit an arc to noisy point clouds. The task of real-time rail recognition needs to take into account the large amount of data, noise of point and various environments of railway. However, these manually designed rail feature processing methods only solve specific problems.

Nowadays, the point cloud processing methods based on deep learning have been developed continuously. They provide a novel way to identify various objects in the point cloud scene. Qi[36] proposed PointNet based on multi-layer perceptron (MLP) for the point cloud tasks of classification and semantic segmentation. The max-pooling layer adapted the disorder of point clouds, but it caused the loss of local details of high-dimensional features. An improved network PointNet++[37] added the hierarchical structure that extracts features from small regions and gradually extends to larger regions. Hu[17] proposed an efficient semantic segmentation method RandLA-Net, which is built on multi random sampling and local feature aggregation module to preserve complex local structures by progressively increasing the receptive field for each point. However, random sampling of these methods discards essential information in the railway, especially for objects with sparse points. 3D voxel partition is an important routine for feature encoding of point clouds. It converts a point cloud into a set of 3D voxels, mainly retaining the 3D geometric information. Huang[38] first divided a point cloud into voxels with the same volume for labeling system. Furthermore, Li[39] designed a 3D backbone network by stacking multiple sparse 3D neural network s to save memory and accelerate computation by using the sparsity of voxels. However, the conventional 3D voxel partition methods obtain limited performance gain because they ignore the sparsity of the point clouds. Zhou[28] employed the cylindrical partition for more robust varying density point clouds.

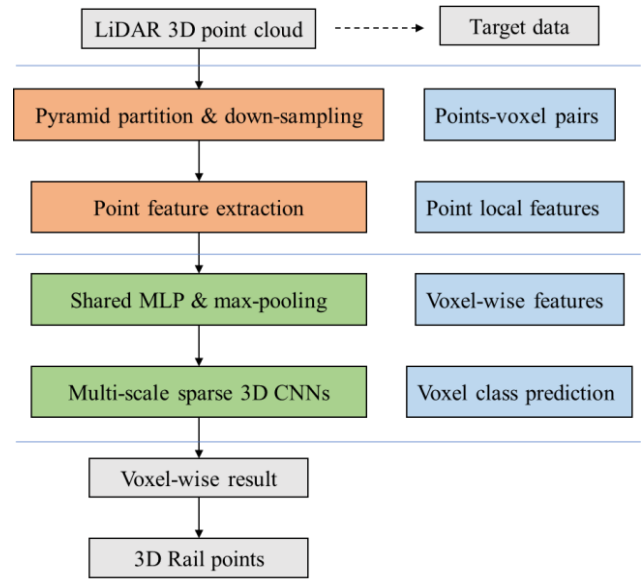

**Figure 1.** The framework of rail recognition based on LiDAR 3D point clouds.

To sum up, the methods based on neural networks improve the accuracy and robustness of point cloud applications. Moreover, the voxel representation naturally preserves the neighborhood structure of 3D point clouds and allows the direct application of 3D CNNs to achieve good performance. However, low resolution of voxel partition results in low accuracy and loss of detail in the object, while high resolution results in high computational costs. An appropriate resolution balances computational cost and acceptable accuracy, both of which are important for rail recognition tasks.

## 3. Method

The method proposed in this study attributes to recognizing rail from 3D point clouds. The framework is shown in Figure 1. It includes four main parts: pre-processing, pyramid partition, features extraction and multi-scale 3D CNNs. The details of

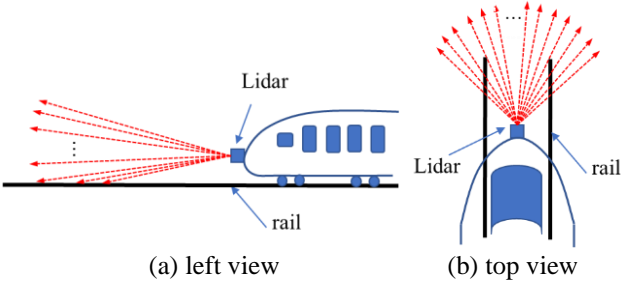

(a) left view　　　　(b) top view
**Figure 2.** The installation position diagram of the LiDAR.

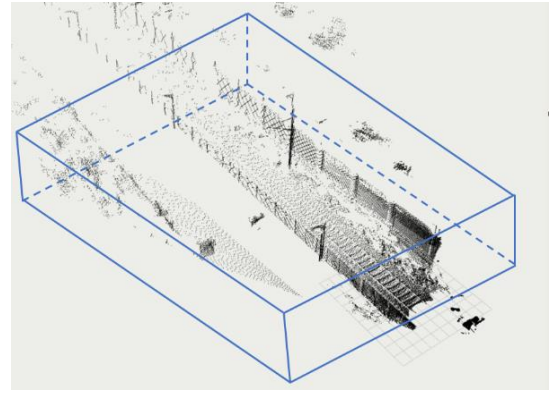

**Figure 3.** A frame point cloud ranged with a cuboid.

the method are given in the following subsections. Firstly, invalid points are filtered during per-processing. Railway 3D point clouds are divided into the points-voxel pairs based on the pyramid partition and down-sampling. Then, the point feature extraction module aggregates neighbor points to obtain the local features. The most salient parts of the rail local features are extracted to produce the voxel-wise features based on shared MLP and max-pooling layers. Finally, the multi-scale sparse 3D CNNs predict the class for each voxel. 3D rail points are segmented from origin point clouds according to the index of the points-voxel pairs.

### 3.1 Point Clouds Pre-processing

The data source of the proposed approach is collected by a solid-state LiDAR which is fixed in front of the train. The pitch and horizontal views of LiDAR scanning are shown in Figure 2, where the dotted lines represent the rays of the LiDAR. In a location with a wide view, the LiDAR can effectively obtain the environmental point clouds data in front of the train.

In the 3D Cartesian coordinate system, taking the LiDAR position as the origin, a frame of the point cloud is represented as follow:

$$F = \{p_n(x_n, y_n, z_n), i_n \mid n = 1, \ldots, N\} \quad (1)$$

where $N$ is the number of points, $p_n$ is the $n$-th point, $x_n$, $y_n$ and $z_n$ are the location of the three dimensions, respectively, and $i_n$ is the LiDAR reflection intensity.

As shown in Figure 3, it is observed that a cuboid ranges a point cloud, and the area out of the cuboid hardly contains any rail points except useless background points for rail recognition. Furthermore, there is much noise outside the region, even beyond the effective scanning range of the LiDAR.

Therefore, these invalid points are deleted in the pre-processing in order to have less effect on rail recognition. Moreover, subsequent calculations reduce the consumption without engaging these points. Let $P$ be a set of points within the inspection area, which can be written as:

$$P = \{p_n(x_n, y_n, z_n), i_n \in F \mid x_{\min} < x_n \leq x_{\max}, \\ y_{\min} < y_n \leq y_{\max}, z_{\min} < z_n \leq z_{\max}\} \quad (2)$$

where $x_{min}$, $x_{max}$, $y_{min}$, $y_{max}$, $z_{min}$ and $z_{max}$ describe the cuboid parameters of the inspection area.

### 3.2 Pyramid Partition and Down Sampling

The solid-state LiDAR is installed at the head of the train and has a fixed viewing range. As shown in Figure 4 (a), the LiDAR scanning area is pyramid-shaped with limited horizontal and pitch angles. Figure 4 (b) and (c) represent a point cloud after pyramid partition and down-sampling, respectively. Due to the factors of distance, occlusion and relative position, the density of the point clouds in the near region is much larger than in the far region. In previous works, the point clouds are divided into cuboid voxels[40], which will lead to the non-equilibrium of the point density of voxels. A new 3D polar coordinate-based pyramid partition method is proposed for increasing voxel size to cover the farther-away region for the railway environment. Voxel volume is positively correlated with the distance range by considering Euclidean distance, pitch and horizontal angle as the resolution. In the pyramid partition, it utilizes the increasing grid size to

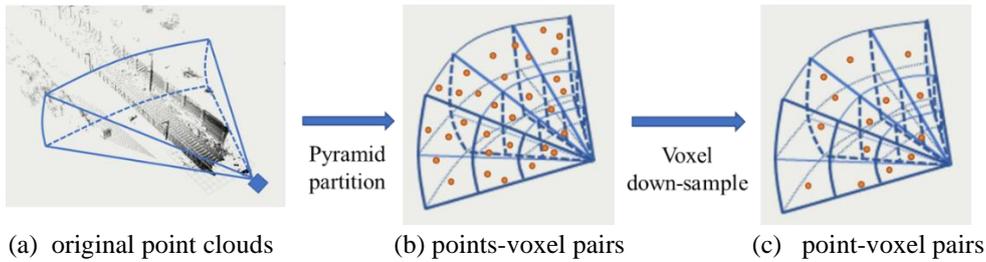

(a) original point clouds　　　(b) points-voxel pairs　　　(c) point-voxel pairs
**Figure 4.** The pyramid partition and voxel down-sample of LiDAR 3D point clouds.

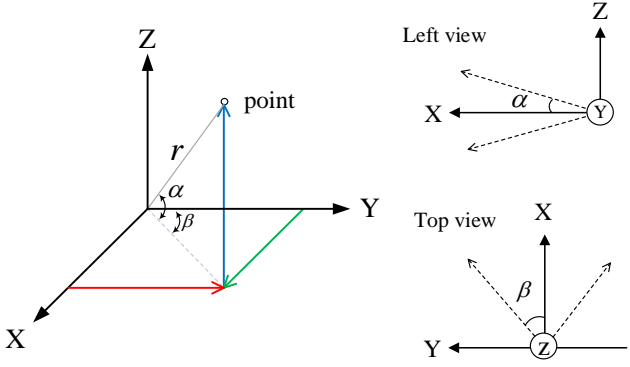

**Figure 5.** The relationship of pitch and horizontal angle for the LiDAR scanning rays in the 3D Cartesian coordinate system. $\alpha$ is the pitch angle, $\beta$ is the horizontal angle, and $r$ is the distance between the point and the origin.

cover the farther-away 3D area. Thus it evenly distributes the points across different voxels and contributes a more balanced representation against varying density and LiDAR scanning distance. The specific steps of our pyramid partition method include coordinate conversion, partition and points grouping.

### 3.2.1 Coordinate Conversion and Partition

As shown in Figure 5, the point clouds in the 3D Cartesian coordinate system are converted into the polar coordinate system according to the angle relationship of the LiDAR scanning rays. The point $p_n(x_n, y_n, z_n)$ represented in the 3D Cartesian system is transformed into the polar coordinate $p_n(\alpha_n, \beta_n, r_n)$, and its corresponding relationship is presented as follows:

$$\begin{aligned} \alpha &= \arctan(z/\sqrt{x^2+y^2}) \\ \beta &= \arctan(y/x) \\ r &= \arctan(\sqrt{x^2+y^2+z^2}) \end{aligned} \quad (3)$$

Given a point cloud, assume the 3D space range be set as $\alpha_{\min}$, $\alpha_{\max}$, $\beta_{\min}$, $\beta_{\max}$, $r_{\min}$ and $r_{\max}$, which represent the minimum and maximum of pitch angle, horizontal angle and distance, respectively. Then the resulting 3D voxel grid is of size $\Delta\alpha$, $\Delta\beta$ and $\Delta r$. Therefore the point cloud is partitioned into voxels, which can be written as:

$$L = \frac{\alpha_{\max}-\alpha_{\min}}{\Delta\alpha}, W = \frac{\beta_{\max}-\beta_{\min}}{\Delta\beta}, H = \frac{r_{\max}-r_{\min}}{\Delta r} \quad (4)$$

where $L$, $W$ and $H$ are the number of voxels in $\alpha$, $\beta$ and $r$ dimensions, respectively.

### 3.2.2 Points Grouping

The points are grouped according to the pyramid voxels that they reside in. Let $V^P$ represent a point cloud consisting of voxels, which can be discretized as:

$$V^P = \left\{ v^P_{l,w,h} \mid \begin{array}{l} l=1,2,\cdots,L \\ w=1,2,\cdots,W \\ h=1,2,\cdots,H \end{array} \right\} \quad (5)$$

where $v^P_{l,w,h}$ is a voxel with many points, $v^P_{l,w,h}$ can be written as:

$$\begin{aligned} v^P_{l,w,h} = \{ p_n(\alpha_n,\beta_n,r_n), i_n \mid [\frac{\alpha_n-\alpha_{\min}}{\Delta\alpha}] = l, \\ [\frac{\beta_n-\beta_{\min}}{\Delta\beta}] = w, [\frac{r_n-r_{\min}}{\Delta r}] = h \} \end{aligned} \quad (6)$$

here, $[\cdot]$ is rounding operation.

The LiDAR 3D point cloud is sparse with variable point density due to factors such as scan distance, occlusion and non-uniform sampling. Thus, a point cloud is partitioned into many voxels containing a variable number of points. Typically, a railway point cloud is composed of 30~90k points. The down-sampling operation remains one point in a voxel. It reduces the redundancy of the point clouds information and the amount of calculation in the point feature extraction part. The voxel-point pair can be written as:

$$v_{l,w,h} = \arg\min(\| v^P_{l,w,h} - v^o_{l,w,h} \|) \quad (7)$$

where $v_{l,w,h}$ represents the point-voxel pair, $v^o_{l,w,h}$ is the current voxel location in the negative direction, and $\|\cdot\|$ is the Euclidean distance between two points. Note that the label must be determined when a voxel includes rail and background points simultaneously. If the number of rail points in the voxel is more than the background points, the label of the current voxel is a rail; otherwise, the label is background.

### 3.3 Point Feature Extraction

A new point feature extraction (PFE) module is proposed for each point to aggregate the geometry information from nearby points. Firstly, the input of the module is a point with the coordinates and feature information, which is down-sampled according to the pyramid partition from the origin point cloud. Then the adjacent points are found from the origin points for the input point based on the k-nearest neighbor (KNN) algorithm. The relative position coordinates are encoded according to the x-y-z coordinates so that each point carries its relative spatial position and geometric features. Otherwise, there is a significant difference in the density of point clouds collected by LiDAR in the railway scene. Thus, the Gaussian density function is introduced as an explicit feature to make each point have a density feature. Finally, the whole network effectively learns the complex local structure in high dimensions through MLP. The module is shown in Figure 6, and the specific process is described in detail below.

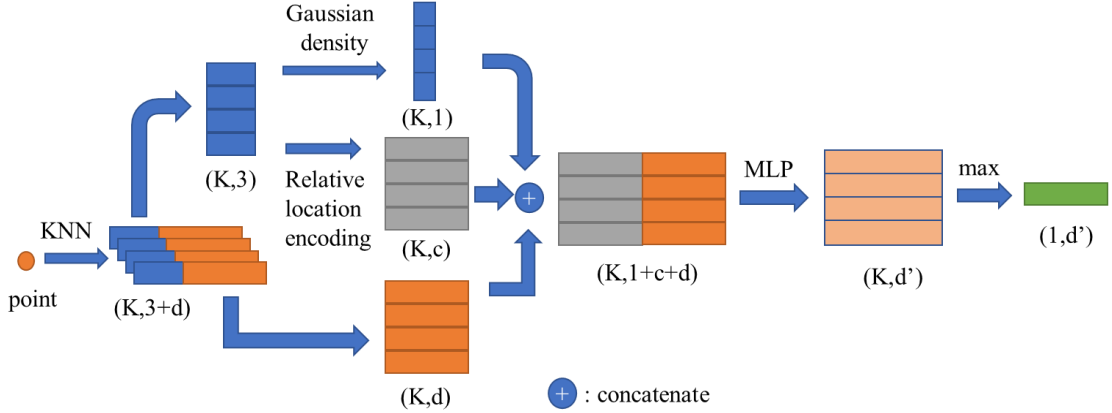

**Figure 6.** Module diagram of point feature extraction.

First, we encode the point locations. A point contains x-y-z coordinates and other features, such as reflection intensity and 3D polar coordinates. The original point cloud are regarded as the support set, and the down-sampled point clouds in section 3.2 are regarded as the query set for the searching operation. For the *n*-th point $p_n$ in a frame point cloud, the nearest K points set $\{p_n^1, p_n^2, \cdots, p_n^k, \cdots, p_n^K\}$ is found based on the KNN algorithm. The relative point location for the nearest *k*-th point is encoded as follows:

$$\mathbf{f}_n^k = (p_n \oplus p_n^k \oplus (p_n^k - p_n) \oplus f_n^k) \tag{8}$$

where $p_n$ and $p_n^k$ are x-y-z coordinates of points, $f_n^k$ is the features of the *k*-th nearest point such as intensity and polar coordinate, and $\oplus$ is the concatenation operation.

Then we consider that the sparse and density variables point clouds are different from the ordered arrangement of pixels in 2D images. The density of point clouds plays a crucial role in railway information. In this paper, the Gaussian distribution probability density function is introduced to determine the density for each point. According to the Euclidean distance of near points obtained by the KNN algorithm, the density between the center point and the nearest *k*-th point is calculated as follow:

$$\rho_n^k = \frac{e^{-r^2/2\sigma^2}}{\sqrt{2\pi}\sigma} \tag{9}$$

where $\rho_n^k$ represent the density of $p_n^k$, $r$ is the Euclidean distance between $p_n$ and $p_n^k$, and $\sigma$ is a hyperparameter as receptive field. Then the density is concatenated to $\mathbf{f}_n^k$:

$$\widehat{\mathbf{f}}_n^k = \rho_n^k \oplus \mathbf{f}_n^k \tag{10}$$

The geometric information of K nearest points is $\widehat{\mathbf{F}}_n = \{\widehat{\mathbf{f}}_n^1, \widehat{\mathbf{f}}_n^2, \cdots, \widehat{\mathbf{f}}_n^k, \cdots \widehat{\mathbf{f}}_n^K\}$. Then point-wise features are obtained from adjacent points based on MLP. Then the most significant features $\mathbf{F}_n$ are extracted from adjacent points by the max-pooling layer, which can be written as:

$$\mathbf{F}_n = \max(MLP(\widehat{\mathbf{F}}_n)) \tag{11}$$

Eventually, the output of the PFE module is a high-rank feature that explicitly encodes the local geometric structures for the center point and the voxel-point pair. After these steps, we define the 3D pyramidic features representation as $\mathbb{R} \in C \times H \times W \times L$, where *C* denotes the feature dimension of the voxel, *H*, *W* and *L* mean pitch angle, horizontal angle and distance, respectively.

### 3.4 Multi-scale Sparse 3D CNNs

In the railway scene point clouds, the rail presents a continuous and equal width strip area, and the beginning of rail is located in the center of the LiDAR view. Inspired by the application of U-net[41] in 2D image semantic segmentation with good performance, a multi-layer residual network structure is designed to fuse the features of point clouds. Figure 7 shows the proposed rail recognition method, where the semantic segmentation network follows the widely used encoder-decoder architecture with skip connections. The encoding and decoding layers include four down-sampling modules and four up-sampling modules based on the block structure of Mobilenet[42]. There are a large number of empty voxels after the partition. Thus, the sparse 3D convolution[43] is used in *L*, *W* and *H* dimensions of point clouds to construct the down-sampling and up-sampling modules. Moreover, recent literature[28] also shows that low-rank kernels build rich high-rank context according to the tensor decomposition theory[44]. In this way, three rank-1 sparse 3D convolution kernels predict the semantic label for each voxel.

## 4. Experiments

In this section, we first introduce the dataset, including point clouds collection, samples and labeling. Secondly, the experiment settings and metrics are provided, and our method tests the real-time performance and accuracy of rail recognition on the dataset. Thirdly, ablation experiments are given to verify the effectiveness of each innovation of the method. Finally, the experiment result report that our work accurately recognizes the rail and achieves real-time performance.

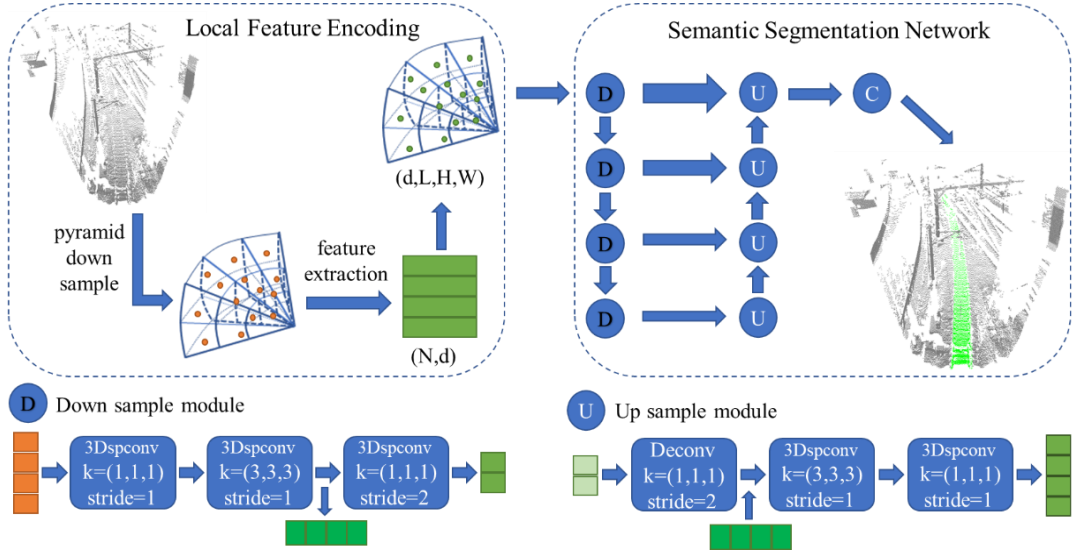

**Figure 7.** The detailed architecture of our semantic segmentation Network. D is the down-sampling module. U is the up-sampling module. C is the sparse 3D convolution layers with three rank-1 kernels. (N, d) represents the number of points and feature dimension, respectively.

*4.1 Data Preparing*

Although there are almost no track scenes in the existing public datasets, we collect the data from China Railway test environment and make the datasets manually for training and validation. The point clouds data of the railway are collected by a Solid-state LiDAR shown in Figure 8, which is mounted at the head of a train with the front view.

The hardware equipment for collecting point clouds is *innovusion Jaguar Gen-1 LiDAR*, and its parameters information is shown in Table 1.

The dataset contains more than 200 million points and 6000 frames. Each point has x-y-z and intensity information, representing the location in the 3D coordinate and the reflection intensity of the object, respectively. According to the characteristics of railway environment and rails, the dataset is classified into six scenes manually, including straight rail, curved rail, intersection rails, multi rails, departure and arrival. The examples of railway 3D point clouds are shown in Figure 9.

The dense point-wise annotation for the rail is labeled manually. The standard rail spacing is 1435mm. The rail area is extended to 1550mm width during labeling in order to obtain the integrity of the rail and the geometric characteristics of the upper and lower drop. In order to reduce dataset redundancy and the workload of manual annotation, continuous point clouds frames are down-sampled to 5 frames per second. Table 2 shows 9 sequences for detailed information of railway data, including point clouds frames, points quantity and rail types. The number of rail points on the railway scene is one-tenth of all railway points. In the open scenes, a frame of the point cloud is composed of about 40k points. However, during the dense scene such as train departure and arrival, the number of points up to 80k.

**Table 1.** the parameters of *Jaguar Gen-1 LiDAR*

| Parameter | Value |
| --- | --- |
| Scanning frequency | 6-10hz, adjustable |
| Pitch angle | 40° |
| Horizontal angle | 65° |
| Range accuracy | <3cm |
| Detection range | 200m |

**Table 2.** The sequences detail information of the railway dataset.

| Seq | Frames | Points quantity (k) All | Points quantity (k) Rail | Rail type |
| --- | --- | --- | --- | --- |
| 1 | 393 | 15,959 | 1,592 | straight, departure |
| 2 | 300 | 12,916 | 1,083 | straight, arrival |
| 3 | 377 | 17,137 | 1,470 | straight |
| 4 | 310 | 9,634 | 1,063 | curve, crossed, multi |
| 5 | 314 | 11,070 | 1,048 | curve, multi, arrival |
| 6 | 310 | 12,005 | 1,083 | curve, crossed |
| 7 | 310 | 10,354 | 1,050 | curve, departure, multi |
| 8 | 308 | 19,367 | 1,195 | curve, departure |
| 9 | 200 | 7,855 | 812 | straight, arrival |

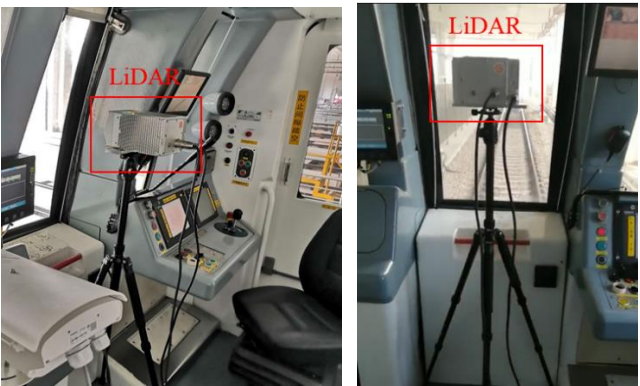

**Figure 8.** LiDAR used in the experiment was mounted at the head of the train.

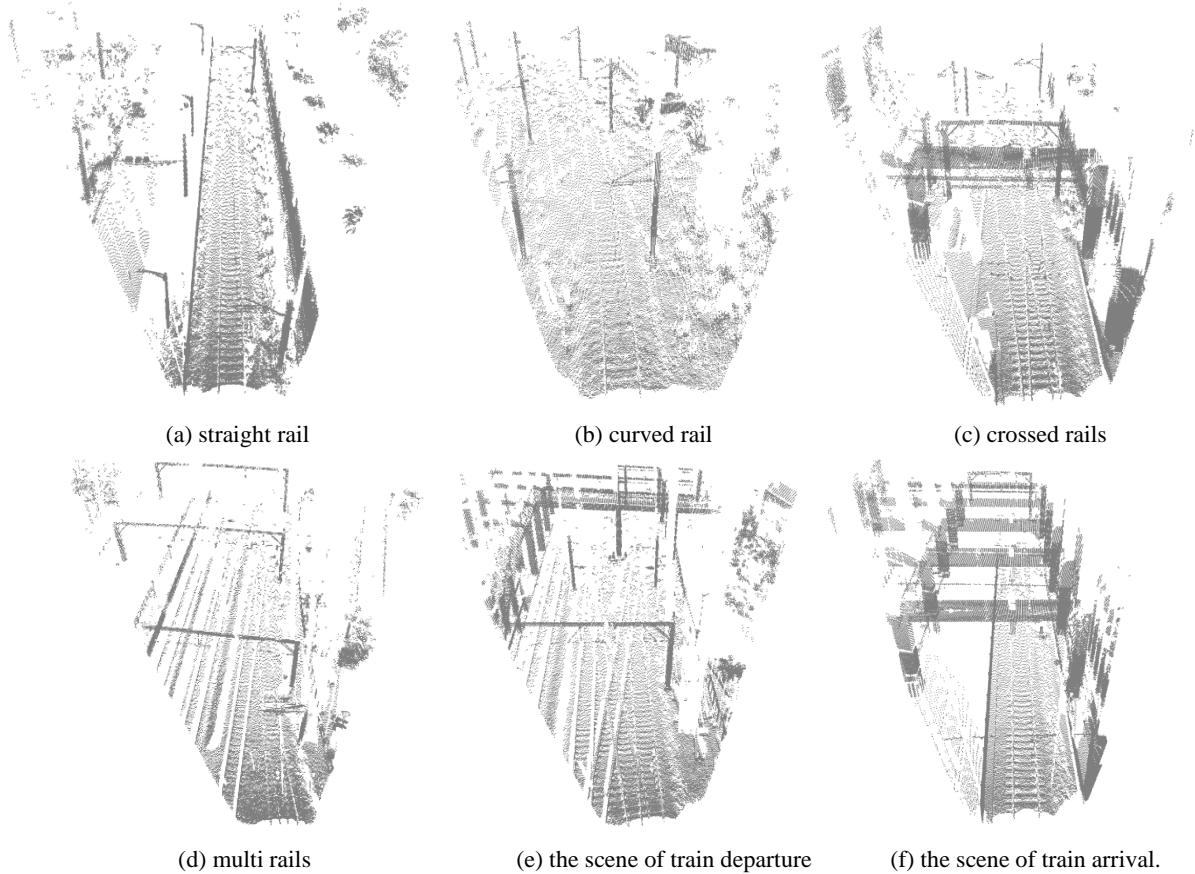

(a) straight rail    (b) curved rail    (c) crossed rails

(d) multi rails    (e) the scene of train departure    (f) the scene of train arrival.

**Figure 9.** Examples for the experiment of rail recognition.

*4.2 Experimental Setup*

All the experiments are performed on a computer with a 6-core 3.4Ghz CPU and a NVIDIA 2080TI GPU. The deep learning environment consists of ubuntu-18.04 system, pytroch-1.6, spconv-1.2.1, etc. The KNN algorithm is implemented in C++, which obviously saves memory.

The setting of major parameters in the pyramid and down-sampling part are listed in Table 3. Then data enhancement is used to reduce the phenomenon of overfitting during the network model training. The point clouds were rotated ± 5° around the z-axis and scaled in three dimensions with ±0.05 times. In addition, Gaussian noise is added to the point clouds. During the model training, 0-6 sequences are regarded as the training dataset. The dynamic learning rate is set with reducing 10 times each 10 epochs, and cross-entropy loss function is introduced to estimate the error.

**Table 3.** Parameters set of the experiments.

| Parameters | Values |
| --- | --- |
| inspection area | [6.5,70,-30,30,-5.5,7] |
| grid size of voxel | [0.3°,0.5°,0.5m] |
| receptive field | 0.5 |
| MLP structure | [13,64,64,16] |
| K | 4 |
| learning rate | 0.001 |
| epoch | 60 |

*4.3 Metrics*

The evaluation metrics employed in our experiments are commonly used methods in point clouds semantic segmentation [10, 17], which are as follow:

$$IoU = \frac{TP}{TP + FP + FN} \quad (12)$$

$$precision = \frac{TP}{TP + FP} \quad (13)$$

$$recall = \frac{TP}{TP + FN} \quad (14)$$

where IoU is Intersection over Union, TP denotes the number of true positive points of rail, FP denotes the number of false positive points that the background points in ground truth are predicted to the rail, FN denotes the number of false negative points that the rail points in ground truth are predicted to the background.

*4.4 Result Evaluation*

After 60 epoch training, the model is verified in sequences 7, 8 and 9, and the results are shown in Table 4. Note that sequence 9 is straight rail, and sequences 7 and 8 are curved rail. The mean IoU of rail recognition is more than 95% for the proposed method.

**Table 4.** The recognition of IoU, precision and recall at testing dataset.

| Sequence | IoU(%) | Precision(%) | Recall(%) |
|---|---|---|---|
| 7 | 96.03 | 98.28 | 97.66 |
| 8 | 93.88 | 96.34 | 97.35 |
| 9 | 95.52 | 98.60 | 96.83 |
| mean | 95.14 | 97.74 | 97.28 |

Besides, the representative semantic segmentation methods Pointnet[36] and RandLA-Net[17] are employed to evaluate the accuracy on our datasets and further compared with our methods. The results are shown in Table 5, which shows that our method achieves the highest IoU, which is 95.14%. Although the time cost and GPU memory usage are performing less than optimally, these are within the acceptable range. The time cost of ours method is more than Pointnet and RandLA-Net, the processing time of the proposed method is 80.9ms which is faster than the collection speed of LiDAR.

**Table 5.** Comparison result of rail recognition.

| Method | IoU(%) | Time (ms) | GPU memory (MB) |
|---|---|---|---|
| Pointnet | 71.88 | 17.6 | 8203 |
| RandLA-Net | 83.42 | 34.5 | 1175 |
| Our method | 95.14 | 80.9 | 1669 |

The visualization samples of rail recognition results are shown in Figure 10, where grey, blue, red and green points represent background point clouds, TP, FN and FP points of rail, respectively. It can be observed that Pointnet only recognizes the straight track well, and the curved and crossed rails show a lot of error points, which means that Pointnet has poor robustness to the complex rail topologies. RandLA-Net uses multi-layer random down-sampling, and it can be seen that the point clouds are obviously sparse than the others. Although the local spatial encoding and attentive pooling modules improve performance in curved and cross rails, there exist a few FP points on both sides of the rail. It is seen that the points of straight rail, curved rail, and crossed rail can be accurately identified by our method, which is the closest to ground truth. The enlarged visualization examples are shown in Figure 11. The blue points are correctly recognized, occupying 95.14% of the rail, which shows good performance of the proposed method. It is observed that there exist few error points in the middle of the rail because the PFE module makes full excavation of the local geometric features of the rail area.

The red and green points indicate the error points which are distributed on both sides and the end of the rail. The point clouds on both sides of the rail are at the junction of subgrade and rail, which contain both rail geometry information and roadbed information. Thus, the points on both sides of the rail are easy to be misidentified. Besides, one of the components in the proposed method is pyramid partition, which may cause points with different categories to be divided into the same cell and bring information loss, especially in the junction of the rail and background.

Due to the scanning range limit of LiDAR, the points at the end of the rail are too sparse to recognize, and the geometric features between the point clouds are very weak. Thus, there are a few crossed rails with the error points.

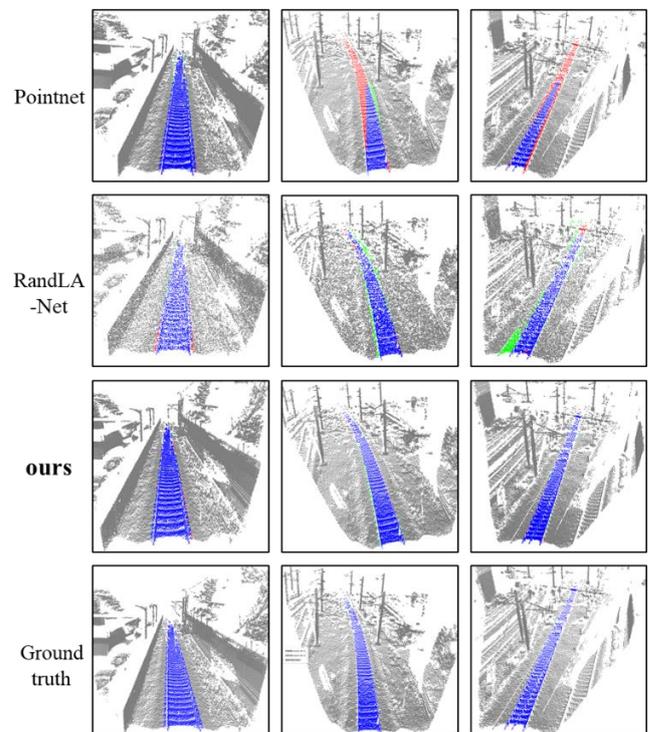

**Figure 10.** Visualization examples of track recognition results.

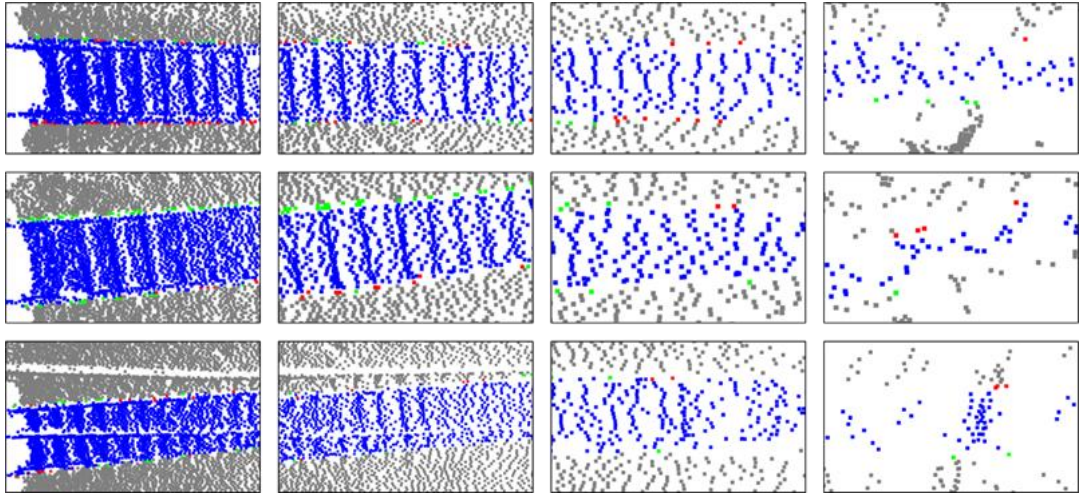

**Figure 11.** The enlarged results of rail recognition which consist of 9 sub-pictures. The sub-pictures on the top, middle and bottom rows correspond to the straight, curved, and crossed rails in Figure 10, respectively. The sub-figures on the left, middle and right columns represent the near, medium and long distances of the rails, respectively.

From Figure 12, it is also observed that the time cost is a positive correlation with the point quantity in a frame point cloud. In the actual railway point clouds collection, the number of points in a frame can reach 70k in the densest scene, such as train arrival and departure, and the number of point clouds is more than 25K in the sparse scene such as single rail condition and plain scene. The experimental results show that the average processing time for a frame of point cloud is 80.9ms. When the train runs at 140km/s, its driving distance during the processing time of the algorithm is 3.2m. The maximum scanning frequency of Jaguar Gen-1 LiDAR is 10hz as shown in Table.1. It means that the collection time for a frame of point cloud by the LiDAR is fixed at 100ms due to the hardware attributes. This indicates that the rail position of the current frame has been recognized when the LiDAR begins to collect the next frame of point cloud. Thus, each frame of point cloud data can be processed on real-time, and our method can satisfy the requirement of real-time performance.

### 4.5 Ablation Study

In this section, additional experiments were conducted to evaluate the effectiveness of each aspect of our method. The results on the railway dataset are reported in Table 6. Our method demotes the algorithm using pyramid partition, PFE module and multi-scale parse 3D convolution networks. Firstly, the pyramid partition is replaced with cube partition to divide railway space into voxels. It is observed that pyramid partition performs better than cube partition with 3.9% IoU gain. Secondly, we remain the integral point clouds without down-sampling. It almost doubles the time consumption and triples the GPU memory usage approximately. Although the IoU raises about 0.6%, it hardly meets the real-time requirements. Thirdly, the PFE module also significantly boosts the performance by about 3% than the MLP, which demonstrates that it is essential to aggregate the local information of point clouds. Moreover, density estimation also improves the recognition accuracy by about 1.1%. To sum up, it can be seen that both the voxel down-sampling based on pyramid partition and PFE module are crucial for rail recognition, which not only improves the performance but also reduces the calculation consumption.

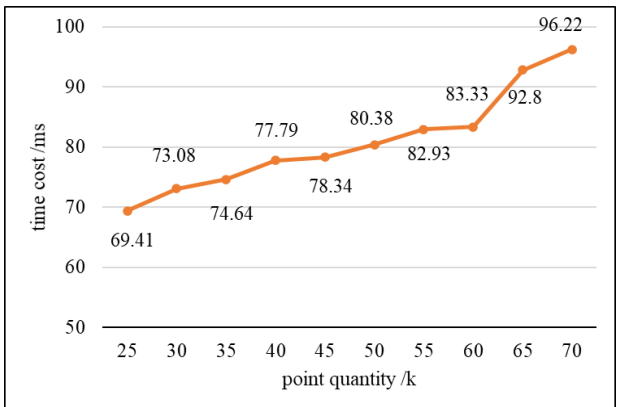

**Figure 12.** Relationship between algorithm time cost and point quantity in a frame point cloud.

**Table 6.** Ablation studies for pyramid partition and PFE module.

|  | IOU (%) | Time (ms) | GPU memory (MB) |
| --- | --- | --- | --- |
| Cube Partition | 91.23 | 73.9 | 1431 |
| Removing down-sampling | 95.75 | 135.6 | 4163 |
| Replacing PFE with MLP | 92.50 | 67.1 | 1403 |
| Removing density | 94.02 | 79.5 | 1607 |
| **Ours method** | 95.14 | 80.9 | 1669 |

## 5. Conclusion

Accurate 3D rail location is a necessary part in the railway support driving system. In this paper, a new semantic segmentation method is proposed for rail recognition based on point clouds. The pyramid partition matches the 3D scanning area of LiDAR, allowing disordered point clouds to be divided into regular voxels. It balances density expression for point clouds. The voxel down-sampling after the pyramid partition saves GPU memory footprint by reducing inputs data volume to the CNNs. The PFE module aggregates the valuable railway information to point-voxel pairs and improves the accuracy of rail recognition. The multi-scale sparse 3D CNNs adaptively identifies railway scenes with multiple topologies. A fast inspection speed (80.9ms per frame) and a high IOU (95.14%) of rail recognition in the experiments that demonstrate the efficiency of the methods.

In future works, the point clouds scanned by LiDAR may cause motion distortion when the train runs at high speed. The time dimension information can be fused into point clouds to solve the motion distortion problem and further increase features for rail recognition. Besides, the data from both spatial and temporal dimensions can also be used to improve rail recognition accuracy. Rail curve fitting and template matching methods may be good ways to predict the long-distance rail location.

## References

[1] J. Wohlfeil, "Vision based rail track and switch recognition for self-localization of trains in a rail network," in *2011 IEEE Intelligent Vehicles Symposium (IV)*, 2011, pp. 1025-1030: IEEE.

[2] M. A. Selver, E. Er, B. Belenlioglu, and Y. Soyaslan, "Camera based driver support system for rail extraction using 2-D Gabor wavelet decompositions and morphological analysis," in *2016 IEEE International Conference on Intelligent Rail Transportation (ICIRT)*, 2016, pp. 270-275: IEEE.

[3] D. Lenior, W. Janssen, M. Neerincx, and K. Schreibers, "Human-factors engineering for smart transport: Decision support for car drivers and train traffic controllers," *Applied ergonomics*, vol. 37, no. 4, pp. 479-490, 2006.

[4] W. Huang, W. Zhang, Y. Du, B. Sun, H. Ma, and F. Li, "Detection of rail corrugation based on fiber laser accelerometers," *Measurement Science and Technology*, vol. 24, no. 9, p. 094014, 2013.

[5] V. Kastrinaki, M. Zervakis, and K. Kalaitzakis, "A survey of video processing techniques for traffic applications," *Image and vision computing*, vol. 21, no. 4, pp. 359-381, 2003.

[6] M. Ruder, N. Mohler, and F. Ahmed, "An obstacle detection system for automated trains," in *IEEE IV2003 Intelligent Vehicles Symposium. Proceedings (Cat. No. 03TH8683)*, 2003, pp. 180-185: IEEE.

[7] F. Maire and A. Bigdeli, "Obstacle-free range determination for rail track maintenance vehicles," in *2010 11th International Conference on Control Automation Robotics & Vision*, 2010, pp. 2172-2178: IEEE.

[8] B. T. Nassu and M. Ukai, "A vision-based approach for rail extraction and its application in a camera pan–tilt control system," *IEEE Transactions on Intelligent Transportation Systems*, vol. 13, no. 4, pp. 1763-1771, 2012.

[9] A. M. Selver, E. Ataç, B. Belenlioglu, S. Dogan, and Y. E. Zoral, "Visual and LIDAR data processing and fusion as an element of real time big data analysis for rail vehicle driver support systems," *Innovative Applications of Big Data in the Railway Industry*, pp. 40-66, 2018.

[10] A. Geiger, P. Lenz, C. Stiller, and R. Urtasun, "Vision meets robotics: The kitti dataset," *The International Journal of Robotics Research*, vol. 32, no. 11, pp. 1231-1237, 2013.

[11] T. Hackel, N. Savinov, L. Ladicky, J. D. Wegner, K. Schindler, and M. Pollefeys, "Semantic3d. net: A new large-scale point cloud classification benchmark," *arXiv preprint arXiv:1704.03847*, 2017.

[12] B. Yi, Y. Yang, Q. Yi, W. Dai, and X. Li, "Novel method for rail wear inspection based on the sparse iterative closest point method," *Measurement Science and Technology*, vol. 28, no. 12, p. 125201, 2017.

[13] M. Taheri Andani, A. Peterson, J. Munoz, and M. Ahmadian, "Railway track irregularity and curvature estimation using doppler LIDAR fiber optics," *Proceedings of the Institution of Mechanical Engineers, Part F: Journal of Rail and Rapid Transit*, vol. 232, no. 1, pp. 63-72, 2018.

[14] M. Arastounia, "Automated recognition of railroad infrastructure in rural areas from LiDAR data," *Remote Sensing*, vol. 7, no. 11, pp. 14916-14938, 2015.

[15] T. Daoust, F. Pomerleau, and T. D. Barfoot, "Light at the end of the tunnel: High-speed lidar-based train localization in challenging underground environments," in *2016 13th Conference on Computer and Robot Vision (CRV)*, 2016, pp. 93-100: IEEE.

[16] E. Nezhadarya, E. Taghavi, R. Razani, B. Liu, and J. Luo, "Adaptive hierarchical down-sampling for point cloud classification," in *Proceedings of the IEEE/CVF Conference on Computer Vision and Pattern Recognition*, 2020, pp. 12956-12964.

[17] Q. Hu *et al.*, "RandLA-Net: Efficient semantic segmentation of large-scale point clouds," in *Proceedings of the IEEE/CVF Conference on Computer Vision and Pattern Recognition*, 2020, pp. 11108-11117.

[18] E. El-Sayed, R. F. Abdel-Kader, H. Nashaat, and M. Marei, "Plane detection in 3D point cloud using octree-balanced density down-sampling and iterative adaptive plane extraction," *IET Image Processing*, vol. 12, no. 9, pp. 1595-1605, 2018.

[19] D. Bolkas and A. Martinez, "Effect of target color and scanning geometry on terrestrial LiDAR point-cloud noise and plane fitting," *Journal of applied geodesy*, vol. 12, no. 1, pp. 109-127, 2018.

[20] Y. Duan, C. Yang, H. Chen, W. Yan, and H. Li, "Low-complexity point cloud denoising for LiDAR by PCA-based dimension reduction," *Optics Communications*, vol. 482, p. 126567, 2021.

[21] Z. Wang *et al.*, "An inverse projective mapping-based approach for robust rail track extraction," in *2015 8th International Congress on Image and Signal Processing (CISP)*, 2015, pp. 888-893: IEEE.

[22] Z. Wang *et al.*, "Geometry constraints-based visual rail track extraction," in *2016 12th World Congress on Intelligent Control and Automation (WCICA)*, 2016, pp. 993-998: IEEE.

[23] K. Duan, S. Bai, L. Xie, H. Qi, Q. Huang, and Q. Tian, "Centernet: Keypoint triplets for object detection," in

*Proceedings of the IEEE/CVF International Conference on Computer Vision*, 2019, pp. 6569-6578.
[24] K. He, G. Gkioxari, P. Dollár, and R. Girshick, "Mask r-cnn," in *Proceedings of the IEEE international conference on computer vision*, 2017, pp. 2961-2969.
[25] D. He *et al.*, "Obstacle detection in the dangerous area of railway track based on convolutional neural network," *Measurement Science and Technology,* 2021.
[26] Z. Wang, X. Wu, G. Yu, and M. Li, "Efficient rail area detection using convolutional neural network," *IEEE Access,* vol. 6, pp. 77656-77664, 2018.
[27] H. Thomas, C. R. Qi, J.-E. Deschaud, B. Marcotegui, F. Goulette, and L. J. Guibas, "Kpconv: Flexible and deformable convolution for point clouds," in *Proceedings of the IEEE/CVF International Conference on Computer Vision*, 2019, pp. 6411-6420.
[28] H. Zhou *et al.*, "Cylinder3d: An effective 3d framework for driving-scene lidar semantic segmentation," *arXiv preprint arXiv:2008.01550,* 2020.
[29] S. Sahebdivani, H. Arefi, and M. Maboudi, "Rail Track Detection and Projection-Based 3D Modeling from UAV Point Cloud," *Sensors,* vol. 20, no. 18, p. 5220, 2020.
[30] C. Chen, T. Zhang, Y. Kan, S. Li, and G. Jin, "A rail extraction algorithm based on the generalized neighborhood height difference from mobile laser scanning data," in *SPIE Future Sensing Technologies*, 2020, vol. 11525, p. 115250N: International Society for Optics and Photonics.
[31] B. Yang and L. Fang, "Automated extraction of 3-D railway tracks from mobile laser scanning point clouds," *IEEE Journal of Selected Topics in Applied Earth Observations and Remote Sensing,* vol. 7, no. 12, pp. 4750-4761, 2014.
[32] K. Qiu, K. Sun, K. Ding, and Z. Shu, "A Fast and Robust Algorithm for Road Edges Extraction from LiDAR Data," *International Archives of the Photogrammetry, Remote Sensing & Spatial Information Sciences,* vol. 41, 2016.
[33] Y. Lou, T. Zhang, J. Tang, W. Song, Y. Zhang, and L. Chen, "A Fast Algorithm for Rail Extraction Using Mobile Laser Scanning Data," *Remote Sensing,* vol. 10, no. 12, p. 1998, 2018.
[34] R. Zou *et al.*, "An Efficient and Accurate Method for Different Configurations Railway Extraction Based on Mobile Laser Scanning," *Remote Sensing,* vol. 11, no. 24, p. 2929, 2019.
[35] S. Huang, C. Ming, S. Lu, S. Chen, and Y. Zha, "A novel algorithm: fitting a spatial arc to noisy point clouds with high accuracy and reproducibility," *Measurement Science and Technology,* 2021.
[36] C. R. Qi, H. Su, K. Mo, and L. J. Guibas, "Pointnet: Deep learning on point sets for 3d classification and segmentation," in *Proceedings of the IEEE conference on computer vision and pattern recognition*, 2017, pp. 652-660.
[37] C. R. Qi, L. Yi, H. Su, and L. J. Guibas, "Pointnet++: Deep hierarchical feature learning on point sets in a metric space," *arXiv preprint arXiv:1706.02413,* 2017.
[38] J. Huang and S. You, "Point cloud labeling using 3d convolutional neural network," in *2016 23rd International Conference on Pattern Recognition (ICPR)*, 2016, pp. 2670-2675: IEEE.
[39] S. Shi *et al.*, "Pv-rcnn: Point-voxel feature set abstraction for 3d object detection," in *Proceedings of the IEEE/CVF Conference on Computer Vision and Pattern Recognition*, 2020, pp. 10529-10538.
[40] Y. Zhou and O. Tuzel, "Voxelnet: End-to-end learning for point cloud based 3d object detection," in *Proceedings of the IEEE Conference on Computer Vision and Pattern Recognition*, 2018, pp. 4490-4499.
[41] O. Ronneberger, P. Fischer, and T. Brox, "U-net: Convolutional networks for biomedical image segmentation," in *International Conference on Medical image computing and computer-assisted intervention*, 2015, pp. 234-241: Springer.
[42] M. Sandler, A. Howard, M. Zhu, A. Zhmoginov, and L.-C. Chen, "Mobilenetv2: Inverted residuals and linear bottlenecks," in *Proceedings of the IEEE conference on computer vision and pattern recognition*, 2018, pp. 4510-4520.
[43] B. Graham, "Sparse 3D convolutional neural networks," *arXiv preprint arXiv:1505.02890,* 2015.
[44] W. Chen *et al.*, "Tensor Low-Rank Reconstruction for Semantic Segmentation," in *European Conference on Computer Vision*, 2020, pp. 52-69: Springer.